\pdfoutput=1
\documentclass{article}

\usepackage{microtype}
\usepackage{graphicx}
\usepackage{subfigure}
\usepackage{tabularx,booktabs}
\usepackage{booktabs} 
\usepackage{caption}
\usepackage{listings}
\usepackage{xcolor}
\usepackage{authblk}

\graphicspath{ {graphics}}
\usepackage{hyperref}

\usepackage[accepted]{mlsys2023}

\def\linkedin{LinkedIn}

\mlsystitlerunning{FLINT: A Platform for Federated Learning Integration}

\begin{document}

\twocolumn[
\mlsystitle{FLINT: A Platform for Federated Learning Integration}



\mlsyssetsymbol{equal}{*}

\begin{mlsysauthorlist}
\mlsysauthor{Ewen Wang}{equal,li}
\mlsysauthor{Ajay Kannan}{li}
\mlsysauthor{Yuefeng Liang}{li}

\mlsysauthor{Boyi Chen}{equal,li}
\mlsysauthor{Mosharaf Chowdhury}{equal,um,consult}
\end{mlsysauthorlist}

\mlsysaffiliation{li}{\linkedin{} Corporation}
\mlsysaffiliation{um}{University of Michigan and RightScope Inc.}
\mlsysaffiliation{consult}{Work done at \linkedin{}}

\mlsyscorrespondingauthor{Ewen Wang}{yuxwang@linkedin.com}
\mlsyscorrespondingauthor{Boyi Chen}{bochen@linkedin.com}

\mlsyskeywords{Machine Learning, MLSys}

\vskip 0.3in

\begin{abstract}
Cross-device federated learning (FL) has been well-studied from algorithmic, system scalability, and training speed perspectives.
Nonetheless, moving from centralized training to cross-device FL for millions or billions of devices presents many risks, including performance loss, developer inertia, poor user experience, and unexpected application failures. 
In addition, the corresponding infrastructure, development costs, and return on investment are difficult to estimate. 
In this paper, we present a device-cloud collaborative FL platform that integrates with an existing machine learning platform, providing tools to measure real-world constraints, assess infrastructure capabilities, evaluate model training performance, and estimate system resource requirements to responsibly bring FL into production.
We also present a \emph{decision workflow} that leverages the FL-integrated platform to comprehensively evaluate the trade-offs of cross-device FL and share our empirical evaluations of business-critical machine learning applications that impact hundreds of millions of users.
\end{abstract}
]



\printAffiliationsAndNotice{\mlsysEqualContribution} 
\section{Introduction}
With increasing computation power and storage capacity in end-user devices, there is a rising trend to move machine learning (ML) toward the edge where data is generated.
One incentive behind this trend is latency reduction in moving computation to the device. 
For instance, for real-time CV and NLP tasks in search and content understanding, sending data in the form of video, audio, or text between user devices and the server is a major bottleneck \cite{lv2022walle}.
Additionally, increasing demands for data protection and privacy in the forms of government regulations (e.g., \citeauthor{GDPR}) and platform restrictions (e.g., App Tracking Transparency from \citeauthor{ATT}) introduce challenges in performing traditional centralized ML on sensitive data. 
These motivate applications like messaging, content recommendation, and advertising, which often rely on potentially sensitive user data to achieve high accuracy, to move ML tasks to the device.

Cross-device federated learning (FL) has captured the zeitgeist as an effective mechanism to address the aforementioned challenges both in industry and academia \cite{kairouz2021advances}.
Federated learning allows distributed ML training on user data on their own devices.
Indeed, federated learning has been successfully deployed on a \emph{case-by-case} basis throughout the industry.
Examples include query suggestions on Google Keyboard \cite{gboard, yang2018applied, chen2019federated, ramaswamy2019federated}, Android smart text selection \cite{smarttext}, applications at Meta \cite{fedbuff, wu2022sustainable}, and several ML tasks on Apple's iOS devices \cite{paulik2021federated}. 
Prior work in cross-device FL has mainly focused on algorithmic improvements \cite{li2020federated, horvath2021fjord}, system scalability \cite{gfl, huba2022papaya}, secure aggregation techniques \cite{so2022lightsecagg}, and model convergence speeds \cite{yu2019parallel}.

However, unlike centralized machine learning, where model architecture and parameters can be tuned and tested in an offline setting, cross-device FL relies on online training systems that require a large population of users to produce utility. 
At \linkedin{}, close to 8,000 types of user devices with more than 150 different OS versions have been observed from use cases in its mobile application.
Having a comprehensive understanding of the impact of different model architectures and hyperparameters on all user devices before deployment is crucial to successful FL training and user satisfaction.
Running resource-intensive ML tasks on user devices can negatively impact user experience and degrade user trust and product experiences.

Building a cross-device FL system and migrating centralized training to that FL system is non-trivial, especially with millions of users.
Forecasting and optimizing infrastructure requirements like resource consumption, data payload restrictions, and model complexity limits for different design choices (e.g., how to collaboratively manage features from cloud and devices, whether to use synchronous or asynchronous training modes and what hyperparameters to use for training) are critical to production success.
A systematic decision workflow to empirically evaluate production design choices is missing in the existing literature.
Many companies have established ML platforms for centralized training, yet do not have the platform and process to estimate the benefits, constraints, and implications of FL. Vice-versa, existing FL platforms today are independent platforms without a close integration with their centralized counterparts.

\textbf{Contributions.} This paper describes the architecture of a novel device-cloud collaborative platform for FL integration, ``FLINT," that augments \linkedin{}'s well-established centralized ML platform. Moreover, it presents a cost-effective decision workflow that leverages the platform to practically assess cross-device FL in \linkedin{}'s context.

In Section~\ref{sec:background}, we describe the traditional ML systems at \linkedin{}, followed by motivations and challenges of cross-device FL. 
Then, Section~\ref{sec:design} describes a detailed FL platform design that closely integrates with the centralized components. This includes an experimental framework that extends the simulation capabilities of FedScale \cite{lai2022fedscale}, which uses device profiles, traces and a virtual clock to provide realistic FL simulations. We have contributed some of our innovations back upstream to the open-source repository.\footnote{https://github.com/SymbioticLab/FedScale}\color{black}
In Section~\ref{sec:eval}, we apply the decision workflow on real use cases at \linkedin{} that could benefit from FL, demonstrating how the FLINT platform can provide ML practitioners with the tools to estimate impact, de-risk projects, and clarify modeling assumptions using a combination of cloud and device resources. 
We show how a close integration with \linkedin{}'s centralized ML platform can help modeling teams evaluate FL in a familiar environment.

Throughout the sections, we share real-world measurements produced by the platform tools, providing insights into an FL system's constraints such as device availability and data/compute heterogeneity. This includes on-device benchmarks of critical, low-latency models on popular device hardware. We demonstrate how these measurements can help forecast model performance under observed real-world constraints and estimate cloud and device resource costs.



\section{Background}
\label{sec:background}





\subsection{Centralized ML at \linkedin{}}
\linkedin{} applies ML to tackle business-critical problems in numerous domains, including advertising, search, messaging, news feed, notifications, and more. Like many traditional ML platforms, a typical ML workflow at \linkedin{} consists of data generation, model training and inference. 
In data generation, data collected in the cloud from multiple sources are anonymized, analyzed, sanity-checked and conflated to extract features and labels for training. 
The data is then used as input for the model training step to perform offline model training and testing. 
The resulting model is deployed by production systems to serve users (both consumer and enterprise). 
At the end, the impressions and actions from users are then logged via tracking events for future data analysis and model iterations. 
For each of these steps, {\linkedin}'s platform uses a combination of reputable open-source and bespoke tools to meet business needs.

\subsection{Motivations for Cross-Device FL}
Increasing demand for privacy and performance is driving the industry to consider moving away from centralized-only ML solutions and instead incorporating computations at the data sources. Similarly, \linkedin{} is considering using cross-device FL for some of its applications. 

\textbf{Privacy and Security.}
The centralized raw data collection and data mixing between users needed to generate training data introduce privacy and security risks. 
With user privacy a priority and a key consideration in \linkedin{}'s product design, there are strong incentives to explore moving some business-critical model training (such as those in the advertising and messaging domains) to user devices to reduce tracking and merging sensitive data.

\textbf{Performance.} 
Moving ML computations closer to their data sources provides better user experiences via improved systems performance.
Many applications (such as search) at \linkedin{} require low latency and high adaptability for recency. 
Models for these applications need to be constantly retrained to adopt the most recent user actions; model inference needs to spend minimal time to deliver predictions. 
Under the centralized model training paradigm, large payloads of user events and inference results have to be transmitted back and forth between devices and the centralized server, which can sometimes introduce significant delays in model freshness and inference speed.

\subsection{Challenges of Cross-Device FL}
Despite the benefits, cross-device FL comes with unique challenges and isn't an end-all solution for all scenarios.
In centralized ML systems, parameter tuning can optimize the model performance assessed by offline evaluations on centralized testing data; system performance can be examined by running trained models using designated hardware; centralized data can be validated, sampled, and shuffled in scalable data pipelines \cite{baylor2017tfx}. 
In contrast, on-device data processing, training, and inference require significantly more careful offline evaluations and system design to responsibly leverage user hardware.
Parameter tuning using user devices can be resource-consuming and deploying faulty or resource-hogging jobs to user devices can harm user trust and negatively impact product reputation and business metrics.

\begin{figure}[ht!]
\includegraphics[width=8.4cm]{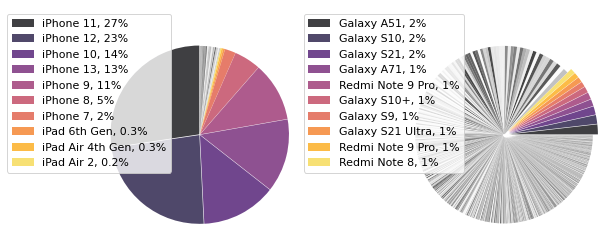}
\caption{Distribution of iOS-based (\textit{left}) and Android-based (\textit{right}) mobile devices in the user base of an example application at \linkedin{}. The gray regions contain device models outside of the legend. Note that Android hardware is much more diverse than iOS hardware, making compute capability challenging to estimate.}
\label{fig:devices}
\end{figure}

\begin{figure}[ht!]
\includegraphics[width=8cm]{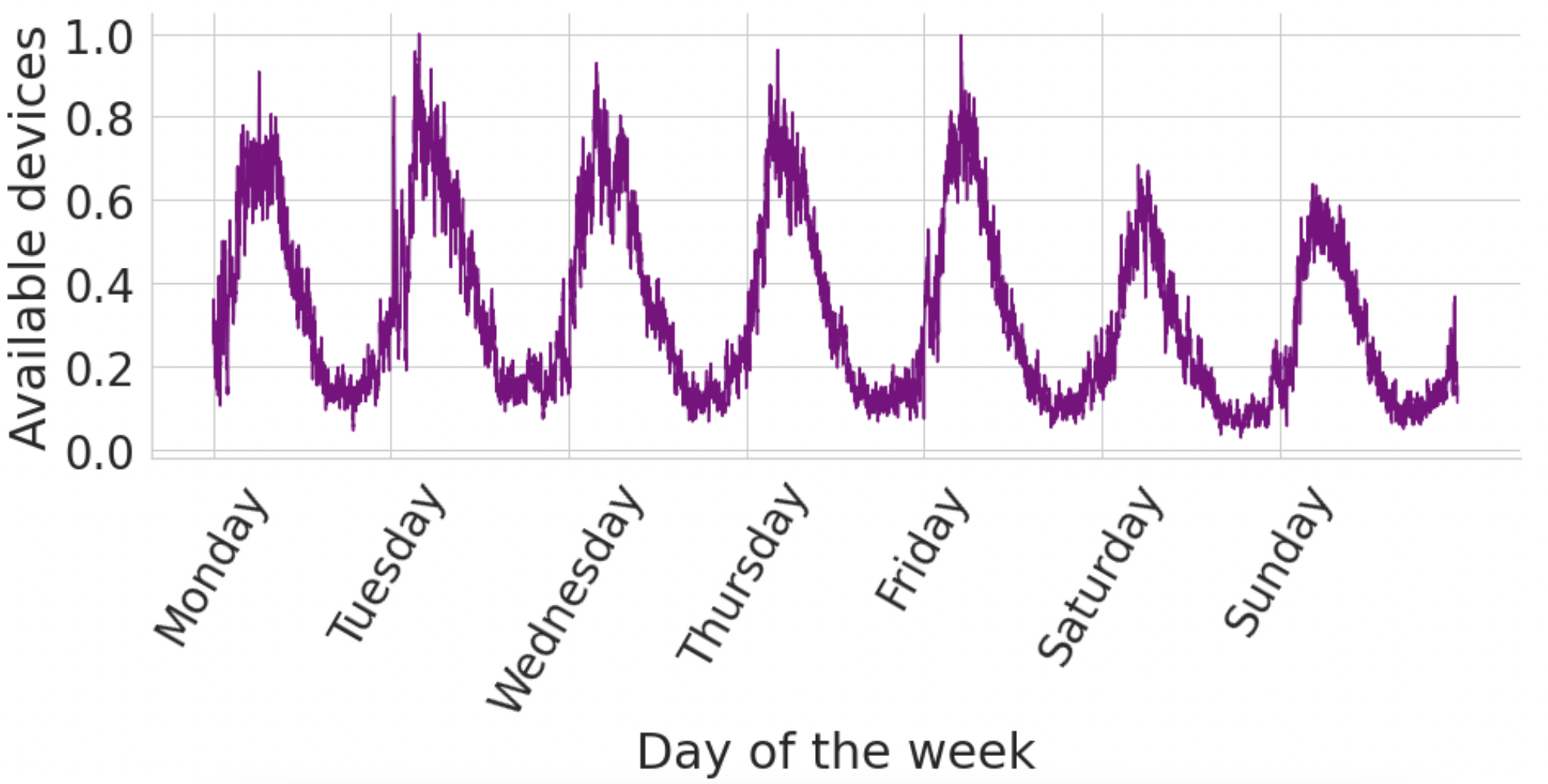}
\caption{Normalized device availability of an example application at \linkedin{} over a one-week period, demonstrating the high fluctuation in client availability.  The predominant trend is that the number of available devices peaks each day and drops to 15\% of the weekly peak at the troughs.}
\label{fig:deviceavil}
\end{figure}

Systems and data heterogeneity present another major challenge. User device heterogeneity (Figure~\ref{fig:devices}) results in significant differences in computation power across various training tasks (Figure~\ref{fig:modeltimes}). Moreover, user behavior differences between devices lead to uneven data availability (Figure~\ref{fig:deviceavil}), diverse data distribution, and violations of feature independence.

These practical challenges put constraints on key aspects of machine learning like model complexity and convergence speed. Ensuring high-quality user experience -- for both the model engineers of the ML platform and users whose devices would participate in training and inference -- requires a comprehensive evaluation framework that considers these system constraints.

\begin{figure*}[h]
  \includegraphics[width=\textwidth]{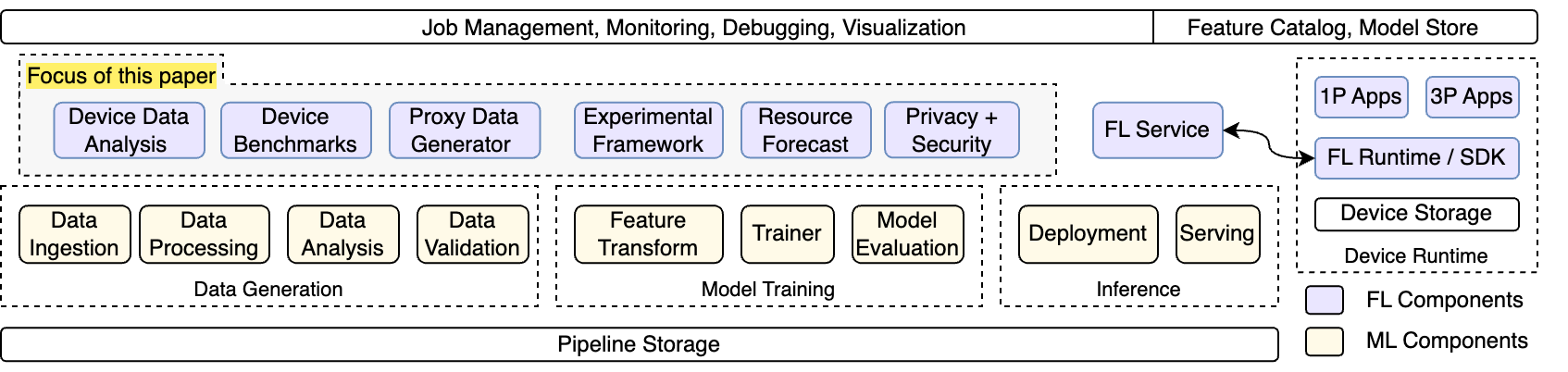}
  \caption{A device-cloud collaborative ML platform with FL integration.}
\label{fig:flplatform}
\end{figure*}

\section{System Design}
\label{sec:design}

\subsection{The Integrated FL Platform at \linkedin{}}
We propose an FL system that works in collaboration with the centralized ML platform (Figure~\ref{fig:flplatform}). It shares common components like model stores, job scheduling, monitoring, and visualization tools with the centralized ML platform described by \cite{baylor2017tfx}, and introduces FL-specific components to enable cross-device FL.

On the device side, an on-device runtime library encodes the FL training and inference tasks and is consumed by 1st party or 3rd party applications.
On the server side, there is an FL server that performs model parameter aggregation and client coordination. The model store, which is shared by centralized training, can store and retrieve versioned parameters during FL training. 
The overall mechanism of the FL server and the device side library works in similar fashions as discussed in other FL system literature \cite{gfl,afl,huba2022papaya,lv2022walle}.

Our proposed FL integration platform, FLINT, builds on top of these well-known FL and centralized ML platforms. This section focuses on 1) tools to leverage centralized data and resources for analyzing FL's impact and viability, 2) a feature catalog that manages both cloud and device-based data, 3) an experimental framework to optimize model performance and system requirements, and 4) a decision workflow that enables decision-makers to understand the constraints, costs, and effectiveness of FL for their business needs.


\subsection{Real World Measurements}
Measuring system constraints from different perspectives helps provide realistic evaluation contexts and guides the design of production systems. 
Running on-device benchmarks before deployment enables engineers to ensure viability of models embedded in heterogeneous software/hardware stacks.
Most existing web services log session metrics and device information during user requests. Our platform tools can analyze this data to produce metrics and visualizations around user device availability patterns and device computation capabilities. 

\begin{figure}[!htbp]
  \includegraphics[width=4cm]{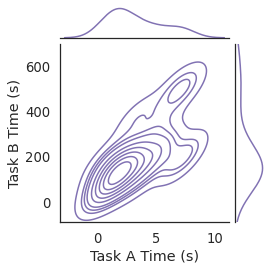}
  \includegraphics[width=4cm]{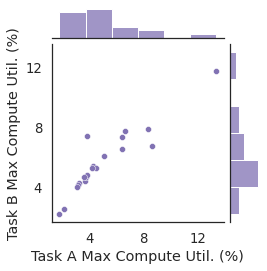}
  \caption{A comparison of two business-critical models' on-device training times and max compute usage percentage over 5,000 examples. This benchmark on 27 device models shows the effects of diverse hardware, and how devices that are optimized for one task might be worse for another. Note the magnitudes difference in training time between FL tasks A and B.}
\label{fig:modeltimes}
\end{figure}

\textbf{On-Device Benchmarks.}
In cross-device FL, the bulk of the computation is offloaded to the clients. Edge devices act as worker nodes in a large computing cluster. 
Importantly, each worker's underlying CPU, GPU, storage, memory, and OS are heterogeneous and could consume vastly different resources to achieve the same task (Figure~\ref{fig:modeltimes}). 
This makes the runtime of an FL task difficult to estimate and could lead to inconsistent user experiences. 
Before allocating such workloads to a heterogeneous population of mobile workers, FLINT's device benchmark step packages models into a benchmark app and deploys it to a pool of test-purposed mobile devices in the cloud, including older and newer generations of popular phones and tablets from Figure \ref{fig:devices}. The collected results (Table~\ref{table:nnarchitectures}) help modelers understand their FL model's worst-case impact on users to derive compatible device models and OS versions for FL participation.

\textbf{User Device Availability.}
We define device availability as pairs of start and end times during which a device can participate in FL training. 
This availability, which fluctuates widely over time (Figure~\ref{fig:deviceavil}), can affect client selection, model fairness, and convergence. 
Our tool helps modelers generate device availability from existing session logs by specifying a set of availability criteria. 
The criteria can include conditions from three categories; 
1) compute capability: based on the device benchmark results, the modeler can generate a list of devices and OS versions that have acceptable worst-case device impact and are compatible with the model architecture; 
2) device state: WiFi connection, battery level, and whether the app is open in the foreground; 
3) user attributes: account reputation, account age, and last participation time, etc. 
These criteria should be iteratively refined to meet the desired model, system, and security needs while ensuring that the model performance is fair among different sub-populations of clients.
For instance, if a device hardware criterion introduces biased model performance on users of older phones, then the hardware requirement needs to be relaxed. And while device charging isn't required for smaller models, a CPU-intensive model (such as Model E in Table~\ref{table:nnarchitectures}), should require a higher battery level ($>$80\%) for participation.
\begin{table}[!htbp]
\caption{Mobile device availability of an example mobile use case at \linkedin{} after applying each participation criteria, showing that only a subset of all users is FL-eligible in practice.}
\label{table:devicefunnel}
\vskip 0.15in
\begin{center}
\begin{small}
\begin{sc}
\scalebox{0.9}{
\begin{tabular}{lccc}
\toprule
Training Criteria & Devices Available \\
\midrule
$A$: Connected to WiFi & 70\% \\
$B$: Battery Level $\geq$ 80\% & 34\% \\
$C$: OS Release $\geq$ Sept. 2019 & 93\% \\
$A \cap B \cap C$ & 22\% \\
\bottomrule
\end{tabular}}
\end{sc}
\end{small}
\end{center}
\vskip -0.15in
\end{table}

In Table~\ref{table:devicefunnel}, we specify a restrictive scenario where conditions $A$, $B$, and $C$ must all be met, leaving only 22\% of the clients available for FL participation.
In this scenario, the training task may require various permissions from the device that may not be available in the background to complete all the sub-tasks (model download/upload, data processing, model evaluation, metrics reporting, etc.). 
This worst-case assumption helps to de-risk potential platform-specific changes to background task permissions. Naturally, app usage duration is tail-heavy and poses a challenge in completing training during short durations of availability.

\subsection{Proxy Data Generator}\label{subsection:proxydatagenerator}
\begin{figure}[!htbp]
\begin{center}
  \includegraphics[width=6.7cm]{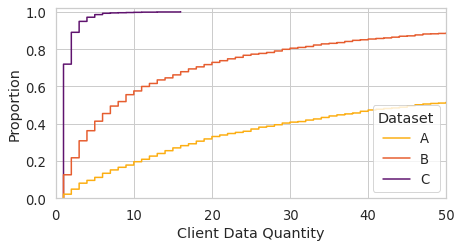}
    \caption{The quantity distribution of key proxy datasets from different domains used in the evaluation, showing that the data sizes between clients in different domains can greatly vary.}
    \label{fig:datasizes}
\end{center}
\end{figure}

To benchmark existing models in FL under realistic heterogeneous conditions, we provide the modeler with a tool to generate per-device proxy datasets from training data in our centralized catalog. In our experiments, the proxy datasets need to be no bigger than centralized training to achieve similar performances. When available, the modeler selects a partitioning field such as obfuscated member or device identifier. The generator uses this field to map records to FL clients. 
When privacy is a concern, the centralized dataset's client-level identifier is discarded.
In these cases, synthetic partitioning strategies \cite{datapartition} can inject label and data quantity skew between the partitions modeled by a Dirichlet distribution. To evaluate the model under varying data heterogeneity, developers can generate multiple versions of a synthetically-split proxy dataset. After generating a proxy, the tool stores it back to the data catalog, adding FL-specific metadata describing feature distributions, client data quantity, label distribution, and client population. These characteristics provide an important understanding of the data heterogeneity between clients (Figure~\ref{fig:datasizes} and Table~\ref{table:datacharacteristics}).

\begin{figure}[!htbp]
  \includegraphics[width=8cm]{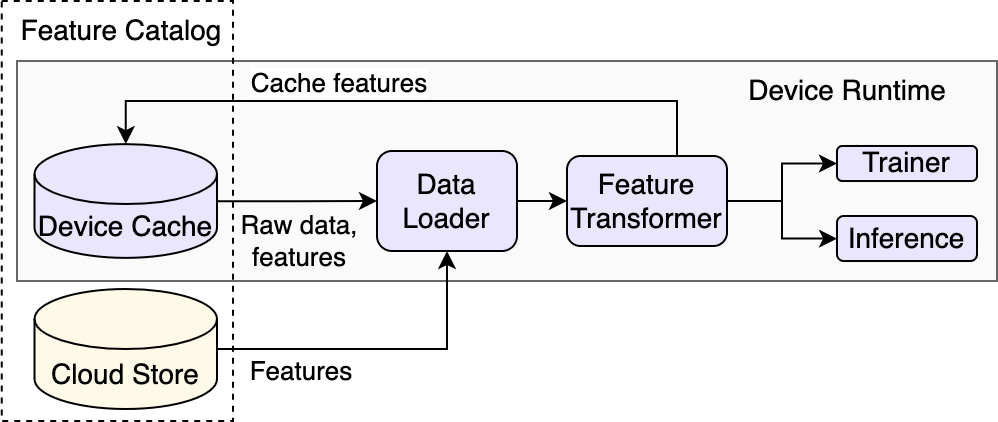}
    \caption{The architecture of a device-cloud feature catalog that manages both device-side and cloud-side features. Certain features and mapping vocabulary can be pulled from the cloud and cached during inference and training. Processed features can also be cached for reuse.}
    \label{fig:datastores}
\end{figure}

\textbf{Data Locality.} Though FL can effectively move compute to the data source, the device runtime should still be able to access cloud-based data through network communication when doing so provides systems and model performance benefits. 
For some tasks, it may be optimal to pull ready-to-use features from the cloud on-demand and join them with device-based contextual features. This reduces the storage and compute footprint of storing and processing large features like embeddings on the device. 
Meanwhile, inference records containing smaller cloud-based features can be cached on the device to reduce network-induced latency during training data processing. 
Additionally, many models require vocabulary files, which contain a set of string to integer ID mappings for features in a dataset, to encode strings into vector values during data processing. The device runtime can pull or cache these files depending on storage usage, WiFi connectivity, and the resource and latency requirements of the task. 
To allow experimenting with combinations of feature management strategies for various applications, FLINT provides a feature catalog (Figure~\ref{fig:datastores}) that manages 1) the device-based features' retention policies and data size limits through cloud-based metadata, 2) the caching strategy of cloud-based features on user devices, and 3) where feature transformations happen. 
The device feature management and caching also allow multiple applications to use overlapping features without duplicated work; when a feature value is created for one task, the runtime can cache it for reuse to reduce latency.

\begin{table}[!htbp]
\caption{Characteristics of sample proxy datasets that are heavily down-sampled on a client level. The max/avg/std values are calculated from client data quantity.}
\label{table:datacharacteristics}
\vskip 0.15in
\begin{center}
\begin{small}
\begin{sc}
\scalebox{0.9}{
\begin{tabular}{lccc}
\toprule
& Dataset A & Dataset B & Dataset C\\
\midrule
Client pop.  & 700,000 & 1,024,950 & 16,422,290 \\
Max records & 39,731 & 103,471 & 406 \\
Avg records & 99 & 184 & 1.53 \\
Std records & 667 & 374 & 1.47 \\
Label ratio & 0.28 & 0.05 & 0.06 \\
Lookback days & 90 & 28 & 61 \\
\bottomrule
\end{tabular}}
\end{sc}
\end{small}
\end{center}
\vskip -0.1in
\end{table}
\subsection{Experimental Framework}
A holistic experimental framework for FL should not only produce model metrics, but also system metrics under realistic constraints.
One goal is to understand the return on investment of FL applications under measured system constraints. 
Another is to predict the infrastructure needs of such a system. 
An added benefit is that modelers can better understand and tune the FL parameters before deploying jobs to devices because offloading all the hyper-parameter tuning workloads to production leads to wasted user resources. 
Our framework builds on top of and significantly extends an open-source FL benchmarking platform to fit our requirements. Deployed on centralized ML clusters, a group of executors poll tasks to run from a leader node, which manages client selection, tracks virtual time, and calculates systems metrics.

\textbf{Inputs and Assumptions}.
In practice, the systems constraints discussed earlier all affect FL's training performance. 
As such, our framework takes multiple real-world inputs to incorporate the complex interactions among these factors in its simulations. 
First, each executor loads a partition of the proxy dataset and maps its records to clients. 
Then, the leader loads and uses device availability records for client selection and task completion decisions.
It then consumes the model's on-device benchmarks (model footprint, processing-time, network usage, etc.), along with the hardware/OS distribution of the users. With these detailed inputs, our framework can report model and system metrics over both virtual clock time and communication rounds to account for data heterogeneity, model complexity, device availability, and hardware capability.

\textbf{Synchronous and Asynchronous Training.}
Our framework supports synchronous FedAvg \cite{fedavg} and asynchronous FedBuff \cite{fedbuff} training modes. 
In practice, client selection is largely dictated by client arrival and availability. Hence, our framework directly selects the next available device from the input sessions at a given virtual time and dispatches a task to an executor. The framework reports results over a virtual time that's calculated independently of the underlying hardware clock. This allows for a better representation of the system in practice when estimating how long a job needs to run, or how much compute time needs to be spent on the device. Even before a client task is dispatched to an executor, the task's duration is calculated using the inputs provided. 
To estimate client $k$'s task duration, we sample $t \gets T$, the distribution of time to train a single example from on-device benchmarks; we also sample a network bandwidth $N$ from Puffer \cite{yan2020learning}, an open-source dataset containing edge device network speeds.
Let $E$ be the number of local epochs, $M$ be the size of a gradient update, and $|D_k|$ be client $k$'s partition size, $taskDuration(k) = t * E * |D_k| + \frac{2 * M}{N}$.

While the asynchronous mode is simpler to implement in a real-time system, it is more difficult to schedule tasks in the right order in a fast-forwarded and distributed simulation. To resolve this, the leader node uses a priority queue-based task scheduler to generate tasks in a streaming fashion and dispatch them to workers in the correct order. From evaluations of different models, we observe that the benefits of an asynchronous system depend on the spread of the client task durations. 
We offer two explanations on why FedBuff \cite{fedbuff} offers faster convergence (Table \ref{table:syncvsasync}): 
1) fewer client tasks have to be started because the aggregation tolerates stale updates, while FedAvg \cite{fedavg} throws away all stragglers; 
2) more client tasks can be started due to the asynchronous task scheduling. 
The effects of 1) and 2) are greater when the client task durations are heavy-tailed and the staleness limit is higher.

\begin{table}[!htbp]\addtolength{\tabcolsep}{-2.5pt}
\caption{Projected training time speedup of FedBuff over FedAvg. The ``client tasks started'' statistic includes failed and stale tasks which are not aggregated. Client computation is the projected sum of processing time on all devices.}
\label{table:syncvsasync}
\vskip 0.15in
\begin{center}
\begin{small}
\begin{sc}
\begin{tabular}{lccc}
\toprule
& Task A & Task B & Task C  \\
\midrule
FedBuff Speed-up & 1.2x & 6x & 2x \\ 
Client Tasks Started & 48.8k & 32.3k & 610k \\
Client Computation & 7.5 hrs & 6.8 days & 25.9 days \\ 
\bottomrule
\end{tabular}
\end{sc}
\end{small}
\end{center}
\vskip -0.1in
\end{table}

\textbf{Scalability and Fault Tolerance.} Using large existing ML clusters and a familiar job management tool, developers can easily simulate millions of clients with our framework. To increase parallelism, a nuance of our proxy data generator tool from Section~\ref{subsection:proxydatagenerator} is that it outputs one partition per \textit{executor} rather than one file per FL client; each partition contains a set of unique clients for an executor to load into memory, which speeds up the random access of client records during training. To support multi-versioned proxies with millions of clients, this strategy prevents an explosion of namespaces on the pipeline storage, which is typically HDFS or cloud blob stores. Furthermore, storing many clients' records together in a file improves the compression ratio.
If each partition still exceeds the memory of the executor, the data can be additionally split by timestamp and swapped in and out during the simulation. This allows a cluster of 20 executors to process over 60,000 client tasks per hour for \textit{Task C} in Table~\ref{table:syncvsasync}; the system scales horizontally and can gracefully handle millions of clients (Table~\ref{table:datacharacteristics}).

For very large experiments, a job could run for days on more than 100 machines. At this scale, the job needs to be fault-tolerant and self-healing. To recover from executor failures, the leader node halts dispatching tasks until all executors have pinged it with a healthy status-code. If a leader node fails, all the executors wait until it is back online to proceed polling for tasks. Since the leader frequently checkpoints the virtual time and recent model weights to the pipeline storage, any restarted leader and executor can resume from the checkpoints without losing more than one round of work.

\begin{figure}[!htbp]
\centering
\includegraphics[width=6cm]{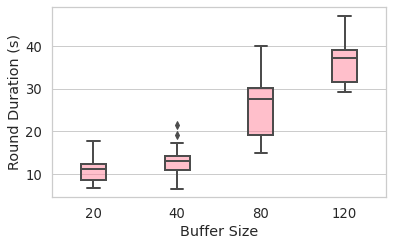}
\caption{Buffer size settings vs time duration to populate the buffer during a sample model's FL training with max concurrency = 180; having a realistic estimation of time during offline evaluation help modelers understand the impact of different parameters.}
\label{fig:buffduration}
\end{figure}

\textbf{Parameter Tuning.} 
An FL system introduces many more parameters to tune, e.g. cohort size, asynchronous buffer size and staleness limits. For example, cohort size is a key parameter that can determine data efficiency and model convergence \cite{charles2021large}, but may have a different optimal value for each application.
However, once a model is deployed, parameter tuning should be done sparingly to responsibly leverage users' device resources.
Additionally, our empirical results (Figure~\ref{fig:lanaupr}) show that model performance under random client sampling can be unstable because clients selected in earlier rounds heavily impact a model's final performance. 
Our experimental framework runs multiple trials of each configuration to report error-bounded metrics. Though such noise can still complicate parameter tuning, parameters selected from proxy datasets can often effectively translate to real FL tasks \cite{kuo2022noisy}. 
Our framework also provides users with an understanding of the relationship between different parameters and model/system metrics.
Figure~\ref{fig:buffduration} shows the relationship between FedBuff's buffer size parameter and estimated round duration.
Figure~\ref{fig:lanaupr} shows how learning rate schedules can affect training stability.

\subsection{Forecasting System Resource} 
\begin{figure}[t]
\includegraphics[width=7.25cm]{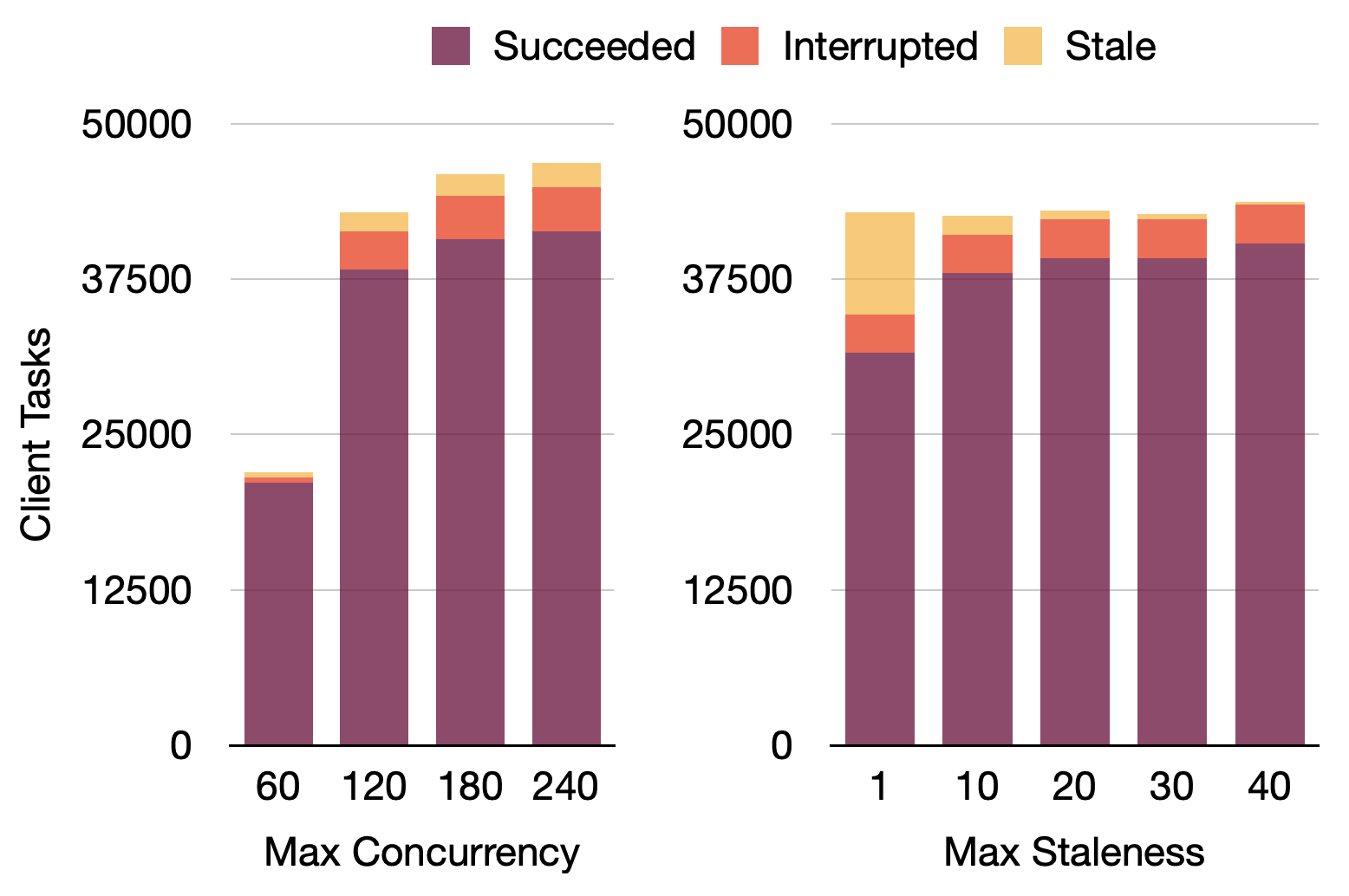}
\caption{Succeeded, interrupted, and stale client tasks under different concurrency and max staleness settings in FedBuff. Higher concurrency can increase both client tasks started and the amount of wasted tasks. Higher staleness tolerance can decrease stale tasks but could slow down learning with older gradients.}
\label{fig:sysmetrics}
\end{figure}

Besides model performance, the FL platform should forecast the overall resource needs from the cloud and user devices, helping engineers optimize the resource efficiency of the system and prevent overloading the finite device and infrastructure capacity.
This can help manage the carbon footprint of edge training jobs, since they can be less energy efficient than centralized training. 
Moreover, renewable energy access at the edge is much more limited due to geographical diversity \cite{wu2022sustainable}.
\\[0.3cm]
\textbf{Reducing Device Resource Consumption.}
In addition to cloud infrastructure costs, a device-cloud platform should account for total edge resource utilization in its notion of budget. As more FL-enabled apps begin sharing the same finite amount of device resources, imposing such a budget can incentivize teams to reduce both cloud and device resource footprint. 
Centralized ML jobs typically specify the workers needed to complete the workload in a reasonable amount of time. While more workers may increase parallelism, it could reduce per-worker utilization, resulting in wasted budget. Similarly in FL, if concurrency is too high, more updates become stale and discarded (see Figure \ref{fig:sysmetrics}). The efficiency of an FL system can be measured with task completion, stragglers, and total device computation time. Our framework reports model performance over these variables so that parameters can be adjusted to reduce the overall user resource footprint. Due to differences in model and data complexity discussed earlier, the sample model for Task C from Table \ref{table:syncvsasync} consumes 620 hours (25.9 days) of client compute time to converge, while the sample model from Task A only takes less than 8 hours. 
The total device time is calculated as
$\sum_{k}^{K}taskDuration(k)$, where $K$ is the sequence of clients that had performed training.
\\[0.5cm]
\textbf{Infrastructure Requirements.} 
Since the trainer in a cross-device FL pipeline is online by nature and handles requests in real-time, a projection of each training job's infrastructure needs is necessary for the modeler to ensure there are enough resources to handle their FL job throughout heavy load swings (Figure~\ref{fig:deviceavil}). 
When multiple FL applications coexist, it is likely for resource contention to occur if they share the same pool of workers for aggregation and coordination. 
The training duration projected by the experimental framework helps to schedule FL workloads efficiently and prevent overloading the service workers due to task overlaps, especially when Trusted Execution Environments (TEE) with limited bandwidth \cite{huba2022papaya} are used for secure aggregation.
In Task C in Table~\ref{table:syncvsasync}, an asynchronous setting where we assume client arrival is uniform, the model takes 48 hours to aggregate 610k tasks (3.53 updates/s).
Multiplied by the size of each gradient update (Table~\ref{table:nnarchitectures}), a TEE needs to receive and aggregate only 2.68MB/second of updates. This demonstrates the framework's ability to project cloud resource needs ahead of deployment based on factors like model size and concurrency.
\color{black}
\begin{figure*}[t]
  \includegraphics[width=\textwidth]{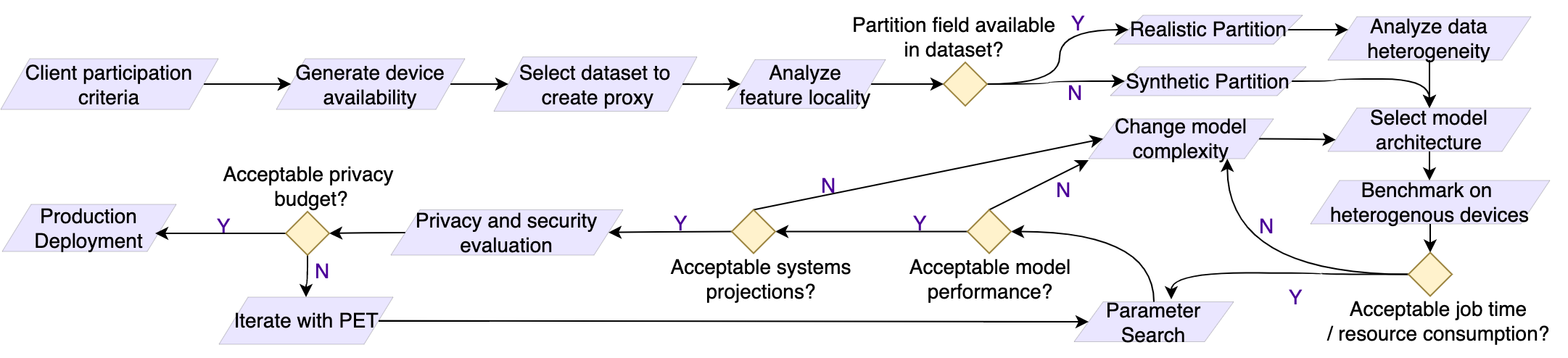}
  \caption{The proposed decision workflow to analyze and bring cross-device FL into production.}
\label{fig:flchecklist}
\end{figure*}
\subsection{Privacy and Security}
Although FL greatly improves user privacy and security by leaving sensitive data on the device, achieving desired privacy properties may still require introducing additional privacy enhancing technologies (PETs) into the system \cite{kairouz2021advances}.  
Currently, developers/security engineers audit the system on a case-by-case basis, since each project has different risk tolerance and privacy budgets.
Our experimental framework can help developers and security experts evaluate the model and resource trade-offs of techniques like FL with differential privacy (FL-DP) \cite{kairouz2021advances}, secure aggregation (SecAgg) \cite{mo2021ppfl}, and robust training \cite{wong2020fast} against adversarial attacks \cite{sun2019can}. Our SecAgg uses TEEs for remote attestation \cite{huba2022papaya}, making it compatible with async FL.\color{black}

\subsection{Decision Workflow}
A standard process to bring FL projects to life at \linkedin{} simplifies many of the production ML operations (MLOps) complexities introduced by FL.
We propose a decision workflow in Figure~\ref{fig:flchecklist} that uses the components of the hybrid FL platform to ensure that the important risks and challenges of each FL project are practically assessed before deployment reaches the users. This covers all aspects of the system, from understanding the client data, compute, and availability, to estimating resource impact, model performance, and privacy/security risks. The process complements the proposed FL/ML platform, leveraging the platform's tools in each of the steps.

\section{Case Studies}
\label{sec:eval}

We apply our decision workflow on three business-critical domains: advertising, messaging, and search. We present the empirical results (Tables~\ref{table:taskresults} and \ref{table:nnarchitectures}), discussing the benefits, systems/performance trade-offs and newfound challenges in the context of the evaluations. Each model is at parity with the centralized model or suffers slight performance loss due to 1) FL's constraints and 2) proxy datasets exclude some features that are only available on-device.

\begin{table}[!htbp]\addtolength{\tabcolsep}{-2.5pt}
\caption{Projected FL training time to reach convergence for each domain's representative model. The performance difference is the median of the FL model's offline metric over N=15 trials compared to the centralized model. We measure ads and messaging performance with Area Under Precision-Recall Curve (AUPR), and search with Normalized Discounted Cumulative Gain (NDCG). In all cases, performance can reach an acceptable range under FL's constraints when compared with centralized training.}
\label{table:taskresults}
\vskip 0.15in
\begin{center}
\begin{small}
\begin{sc}
\begin{tabular}{lllll}
\toprule
 & Ads & Messaging & Search \\
\midrule
Training Time &  4.2 days & 18.9 hrs & 2.58 hrs  \\
Performance Diff. &  -1.85\%  & -0.18\% &  -1.64\% \\ 
\bottomrule
\end{tabular}
\end{sc}
\end{small}
\end{center}
\vskip -0.1in
\end{table}

\begin{table*}[ht]\addtolength{\tabcolsep}{-2.9pt}
  \caption{On-device evaluation of device-capable model architectures selected to represent common ML tasks at \linkedin{}. We report mean training times and CPU utilization \% for each model over 5,000 records, aggregated across 27 devices with diverse hardware.}
  \label{table:nnarchitectures}
  \centering
  \begin{tabularx}{\textwidth}{p{1.2cm}p{4.4cm}*{9}{X}p{1.2cm}}
    \toprule
    Model & Description & Trainable Params & Storage (MB) & Network (MB) & Memory (MB) & Mean Time (s) &     Stdev Time (s) & Mean CPU (\%)\\
    \midrule
    \textit{A} & Tiny Neural Net  & 1.51k & 0.057 & 0.11 & 3.08  & 4.98 & 3.37 & 1.63 \\
    \textit{B} & MLP w/ sparse features  & 189k & 0.76 & 1.52 & 10.64 & 61.81 & 44.17 & 3.91\\
    \textit{C} & MLP w/ medium embedding & 208k & 0.85 & 1.88 & 0.85  & 3.26 & 2.23 & 5.29\\
    \textit{D} & CNN w/ large embedding & 390k & 10.79 & 3.12 & 8.37 & 70.13 & 50.82 & 4.72\\
    \textit{E} & Multi-task MLP & 922k & 7.52 & 7.38 & 43.14  & 238.38 & 178.13 & 6.43\\
    \bottomrule
  \end{tabularx}
\end{table*}

\subsection{Advertising}

Privacy in machine learning has received significant attention in recent years. 
Traditional machine learning in the digital advertising industry relies on collecting user data for measurement, targeting, and click/conversion predictions.
By reducing sensitive data tracking, cross-device FL enables private model training that can improve advertising quality, member trust and safety. 
In this section, we describe the detailed steps we took to evaluate FL on an advertising use case, and the results that demonstrate the potential of moving to cross-device FL while revealing several practical challenges. 

\textbf{Client Participation and Availability}. First, we define our client participation criteria: (a) app is open in the foreground, (b) battery level $>$ 80\%, and (c) connected to WiFi.
Our criteria are designed to be conservative, choosing to err on the side of classifying a device as not FL-ready when in doubt.
For example, we require (a) because if for some reason CPU or battery usage spikes when the app is in the background, a phone OS could choose to kill our training process.  To eliminate this possibility altogether, we do not count any app background time as time we can use for FL. We then use these filters to generate device availability traces.
We query for two weeks of anonymized session data from the LinkedIn app, since usage tends to exhibit weekly periodicity.
Short gaps where the app is in the background are subtracted from the availability session duration, whereas longer gaps split a session into two.
Since we only have battery level and WiFi connectivity data for a smaller subset of mobile usage, we calculate empirical probabilities of WiFi connection and high battery level over time (Table \ref{table:devicefunnel}). For each session from our query, we perform a weighted coin-flip based on the session's start time to decide whether to include or exclude it from the output device traces.

\textbf{Building a Proxy Dataset.} Next, we use a centralized dataset in advertising that is down-sampled on a client level to preserve the natural quantity and label skew. We then analyze the feature locality of the data to move it into an on-device setting. In this domain, a candidate is typically a potential advertisement to display or an targeting-segment that is scored in the context of the user. This application retrieves 184 candidates in a single request on average from the server, which includes some server-side features. Afterwards, each candidate is decorated with client-side features and similarity scores are calculated. To create a proxy dataset, we create a client id field based on the member id, and map each unique id to an integer for further anonymization. 
Then, we run a Spark job that groups the examples by client ids and computes inter-client statistics. Through this analysis, we find that client data is non-IID and extremely tail-heavy due to users engaging disproportionately more on the app (std. of 667, and max of 39,731 records). 

\textbf{Selecting a Mobile-ready Model.} Next, we analyze three model architectures that are tested in a centralized setting. Models that need to be deployed in third-party apps via an SDK have stricter size requirements ($<$1MB), while critical models in the first-party app have looser storage constraints ($<$10MB). Thoroughly evaluating resource footprint requires taking measurements on device, since model complexity alone is not a good predictor (compare the resource footprints between Models \textit{A}, \textit{B}, and \textit{C} in Table \ref{table:nnarchitectures}). We convert our three candidate models to a TFLite format, and deploy them for training on dummy data to 27 different devices on AWS Device Farm in our benchmarking app. Out of the three architectures, we picked the model that satisfies the size requirement mentioned earlier at 0.76MB and consumed the least network and memory. This also helps us validate that the ops bundled with the ML runtime are sufficient to execute the model training.

While we observe that the model training footprint is acceptable, model assets may pose a challenge. During data processing, the feature transformer must map more than 70\% of its features from categorical values to unique indices through vocabulary files during the data pre-processing step. 
Though these mappings work well in a centralized setting, the device must refresh and store vocab files as assets, which could be as big as 1.28MB for high-cardinality variables.
To overcome the memory and disk constraints, feature hashing \cite{featurehash} can perform the mapping through a hash function, trading less storage space with lower predictive power (due to hash collisions).

\begin{figure}[!htbp]
\begin{center}
\includegraphics[width=7cm]{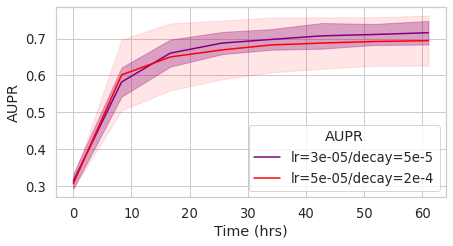}
\caption{AUPR of an example model trained using two different exponential decay LR schedules on N=5 trials each. This shows a good learning rate schedule can improve training stability.}
\label{fig:lanaupr}
\end{center}
\end{figure}

\textbf{Systems and Model Performance.} Next, we partition the proxy dataset for 20 workers by client id in a round-robin fashion to enable faster job execution time in a cluster. This number is picked roughly based on the total data size divided by the memory available per worker. Our job config specifies the device traces, on-device performance distributions produced earlier, and other hyper-parameters to realistically evaluate the FL training. Shown in Figure~\ref{fig:lanaupr}, model performance under random client sampling can be highly variable due to data heterogeneity, as the clients selected in the earlier rounds can determine the model's convergence. We use such experiments to tune parameters (such as the learning rate schedule in Figure~\ref{fig:lanaupr}) before production deployment. With the scalability of the framework, we repeat each trial 5 times to estimate an error bound. We decide that the projected training time of 4.2 days is an acceptable SLA when FedBuff async training is enabled, as the centralized counterpart only needs to be retrained weekly. The performance difference in Table~\ref{table:taskresults} is also acceptable; since this on-device deployment helps meet critical compliance and regulation requirements in the ads industry, there is a higher tolerance for accuracy degradation (up to 5\%). Moreover, the proxy dataset is only a subset of all the signals that can be consumed on device; hence it's a worst-case estimation.

\textbf{Security and Privacy.} Transitioning from a centralized setting where signals are collected, the data minimization already greatly improves the product's privacy budget without any additional PETs. Nonetheless, we project the data transfer bandwidth needed from a TEE is under 3MB/s, which is sufficiently within the limit. From the security evaluation of this case study, in which the model is distributed via an SDK, we identify a new attack scenario if it is possible for the SDK's host application to control a significant portion of the FL participants, hence poisoning the data or updates of a group of clients. This unique hub-and-spoke setup prompts further security research on detection and defenses.

\subsection{Messaging}
Consumer messaging applications often contain highly confidential data and are encrypted end-to-end. This poses restrictions on the data that on-premise ML tasks like abuse detection and smart-inbox features could use. Cross-device FL enables message data to be used for training in its original state on the device. To create a proxy dataset without data decryption, we partition a dataset of synthetic messages used for centralized training. The FL training achieves a promising performance compared to the centralized training, with only a 0.18\% difference in the test metrics. This difference is negligible given the improved freshness of the training data, which helps the global model quickly adapt to user feedback. Lastly, the evaluation process helps us identify practical on-device challenges in this domain.

\textbf{Size of Text Embeddings.} 
Large mobile apps can discourage downloads and increase uninstalls. 
Many deep NLP models contain word embedding tables to map text tokens into fixed-size embeddings that are fed into the rest of the layers. One of our centralized models in the messaging domain initially has a 150 million parameter embedding layer greater than 500MB, prohibiting on-device deployment.
Reducing the vocabulary from 500K words to 50k and the embedding dimension from 300 to 50 leads to a 60-fold size decrease, fitting the 10MB size constraint. Other solutions include embedding compression methods like TT-Rec \cite{yin2021tt} or MEmCom \cite{memcom}. Finally, the application can bundle a text embedding that's shared by NLP models in different domains (search, recommendations, etc.), and download a smaller language-specific subset of the corpus based on the user's language. 

\textbf{Security.} 
Evasion attacks involve adversaries carefully crafting samples fed into the model to change the inference result, presenting a practical concern for message abuse and scam detection models during inference time. 
Access to the model is especially a concern if a bad actor could decrypt the weights stored on the device. Existing defenses involve robust training, but generating adversarial examples during training can be expensive, \cite{wong2020fast, hao2022tale} even more so on the device. This introduces a tradeoff between model robustness and resource consumption. 
Data poisoning attacks are another concern when enough users coordinate to generate fake messages and corresponding actions, though this usually requires adversaries controlling an impractical portion of the population \cite{shejwalkar2022back}. 
Using the FLINT platform, our decision workflow enables evaluating new mitigation strategies; for instance, a more robust client selection criteria that incorporates the user's reputation score and account age, or continuous FL training to adapt recent user feedback.

\subsection{Search}
In industry, FL has been used in browser URL bar suggestions by locally training on private browsing history \cite{hartmann2019federated} and ranking keyboard suggestions \cite{gboard}. 
Naturally, training ranking tasks on device allows directly using the displayed candidates and user feedback to generate training data directly on the device.
At \linkedin{}, most production search workflows are bounded by strict latency budgets in the sub-100ms range \cite{guo2021deep}. Query autosuggestion and completion require instant predictions to feel responsive; search ranking models need regular retraining to reflect search trends. 
On-device ML has the potential to improve model freshness and reduce inference latency. 
In ranking tasks, the application can locally cache, retrieve, and rank frequent documents without any network communication. For language generation tasks like query completion, locally-trained LSTMs can generate more personalized search suggestions using partial queries.

Our evaluation of a low-latency model in the search domain shows a performance difference of only 1.64\% (Table~\ref{table:taskresults}) when trained on FL under realistic system constraints, with minimal device resource usage.
Moreover, FL training can reduce the resources needed to store/ETL data and regularly retrain models in data centers. 
However, similar to advertising, training data in search can have a very high quantity skew because of ``superusers". 

\section{Related Work}

\textbf{Systems for Federated Learning.}
Several large-scale cross-device FL systems have been proposed in recent years, most notably Google's GFL system \cite{gfl}, Apple's FL system \cite{afl}, and Meta's PAPAYA \cite{huba2022papaya}.
The designs of our service and client run-time draw inspiration from all of them. Our design echoes PAPAYA by supporting both sync and async, selecting clients based on demand by active tasks.
Our sync mode is similar to GFL's round-based design and uses client over-commitment to handle dropouts.

\textbf{Evaluation Frameworks.}
To build the experimental framework, which to the best of our knowledge is first described in this paper, we considered many existing open-source FL toolkits that provide simulation capability, e.g., TFF \cite{bonawitztensorflow}, FLUTE \cite{dimitriadis2022flute}, Flower \cite{beutel2022flower}, and FedML \cite{he2020fedml}. 
While they provide a variety of models, datasets, and algorithms for benchmarks, they report results over communication rounds. 
Our design expands on FedScale \cite{lai2022fedscale}, reporting both model and system metrics reported over a virtual time and communication rounds to account for model complexity, device availability, compute and network, etc. 

\textbf{FL Benchmarks.} 
Many popular cross-device FL benchmarks \cite{caldas2018leaf} are focused on CV and NLP tasks (FEMNIST, CIFAR10, Reddit, Shakespeare etc.), and have helped drive FL research and algorithmic improvements in the recent years. In general, our work prompts the design of more tabular FL datasets with sparse features, noisy and imbalanced labels, and heavy data quantity skew.
This is representative of in-app user behavior in the wild, where data is often scarce and noisy, and superusers dominate. An equally important consideration for realistic benchmarks is whether the models that are benchmarked can be deployed to lower-end user devices (and be small enough to co-exist with other models in a mobile app).
We suggest researchers report a measure of model size and on-device resource usage with the benchmarks of FL models.

The open-source benchmarks implemented in FedScale \cite{lai2022fedscale} are close proxies for our case studies given the naturally-partitioned datasets. The Taobao Ad Display / Click dataset (label ratio: 5.4\%, clients: 1.1 mil, mean: 23, std: 65) is a good proxy for our advertising scenario because it captures the scarcity of user response. Models B and C from Table~\ref{table:nnarchitectures} fit in the ballpark in terms of architecture, size and performance requirements. For message classification tasks, the Amazon Review dataset is a good proxy (clients: 256,059, mean: 2.2, std: 4.4). Section~4.2 describes the model architecture deployed in a typical message classification task. As for next-word query prediction in search, we believe there are already mature benchmarks such as Stackoverflow and Reddit, modeled by on-device LSTMs \cite{hard2018federated}. In search ranking, there is a gap for a federated learning-to-rank dataset with a natural partitioning. Lastly, FedScale incorporates device availability traces from \cite{yang2021characterizing}, which captures a similar weekly fluctuation pattern with a difference of 4x between peak and low, given the device is plugged-in and idle. Our availability in Figure~\ref{fig:deviceavil} fluctuates by a factor of 14x due to strict participation requirements and geographically-based usage patterns, serving as an upper-bound. The device traces an be re-sampled and adjusted based on the deployment scenario.
\color{black}

\textbf{ML Platforms.} 
The learning algorithm itself is only one component of an ML platform \cite{sparks2017keystoneml}. Systematically deploying ML in production has received large attention over the past decade with many available MLOps and platform solutions \cite{baylor2017tfx, zaharia2018accelerating} because gluing together disjoint components may do a job once, but often leads to significant technical debt \cite{hidden-debt}. With increasing data and model sizes, many parallelism techniques require the orchestration of distributed systems \cite{sergeev2018horovod, moritz2018ray}. 
Lastly, ML platforms need to be user-friendly via automation and declarative solutions so that even non-experts can leverage ML \cite{kraska2013mlbase}.

\textbf{FL Platforms.} The concept of device-cloud collaborative ML platforms is not new. Alibaba's Wall-E \cite{lv2022walle} provides a deployment platform and high-performance mobile compute runtime for on-device tasks. To the best of our knowledge, FLINT is the first to fill the important gaps to allow an effective coexistence of centralized and on-device ML applications. We believe such a platform should perform all the tasks of an ML platform while providing the tools to analyze and make decisions based on the systems and data challenges inherent to FL.


\section{Concluding Remarks}
As shown by our evaluations of three business-critical ML applications, a cloud-device collaborative FL platform can help ML developers and decision makers practically assess the systems constraints, costs, and benefits of production FL projects. 
Leveraging the platform, a systematic decision workflow can help teams responsibly bring FL projects to hundreds of millions of users at \linkedin{}. 
Our results also confirm that in industry scenarios where users could benefit from improved system performance and data privacy, FL has the potential to replace centralized training.

In literature, most cross-device FL benchmarks and systems are designed to process purely device-generated data (text, voice, image), and their components operate in standalone FL platforms.
As shown in our practical scenarios, the model/system performance and user experience in FL can greatly benefit from a collaboration of device-side and cloud-side data and systems. 
Hence we emphasize further innovations in the device-cloud platform space.

\section*{Acknowledgements}
We would like to thank 
Lu An, 
Sudhanshu Arora, 
Oscar Bonilla, 
Ting Chen, 
Alexey Dubovkin,
Ebrahim Emami, 
Humberto Gonzalez,
Ankit Goyal,
Mingyang Hu, 
Abelino Jimenez, 
Raghavan Muthuregunathan,
Haowen Ning,
Ray Ortigas,
Yafei Wang,
YuanKun Xue,
Hao Yu, 
Leighton Zhang,  
Haifeng Zhao,
and Tong Zhou for their valuable feedback on this work. Further, we thank Siyao Sun, Rahul Tandra, Zheng Li and Souvik Ghosh for their continuous support throughout this project.



\nocite{langley00}

\bibliography{paper}
\bibliographystyle{mlsys2023}

\end{document}